\documentclass[10pt,twocolumn,letterpaper]{article}

\usepackage[pagenumbers]{cvpr} % To force page numbers, e.g. for an arXiv version

\definecolor{cvprblue}{rgb}{0.21,0.49,0.74}
\usepackage[pagebackref,breaklinks,colorlinks,allcolors=cvprblue]{hyperref}

\def\paperID{969} % *** Enter the Paper ID here
\def\confName{CVPR}
\def\confYear{2025}

\title{Human Multi-View Synthesis from a Single-View Model: \\Transferred Body and Face Representations}

\author{Yu Feng\textsuperscript{1}, Shunsi Zhang\textsuperscript{2}, Jian Shu\textsuperscript{1}, Hanfeng Zhao\textsuperscript{2}, \\ 
Guoliang Pang\textsuperscript{2}, Chi Zhang\textsuperscript{3}, Hao Wang\textsuperscript{1} \\
\textsuperscript{1}\textit{The Hong Kong University of Science and Technology (Guangzhou)}, China \\
\textsuperscript{2}\textit{Guangzhou Quwan Network Technology}, China \\
\textsuperscript{3}\textit{Westlake University}, China \\
Email: yufeng9819@gmail.com, haowang@hkust-gz.edu.cn \\
}
\begin{document}

% \maketitle

\twocolumn[{%
\maketitle
\begin{center}
\vspace{-0.1in}
    \centering
    \captionsetup{type=figure}
    \includegraphics[width=0.99\textwidth]{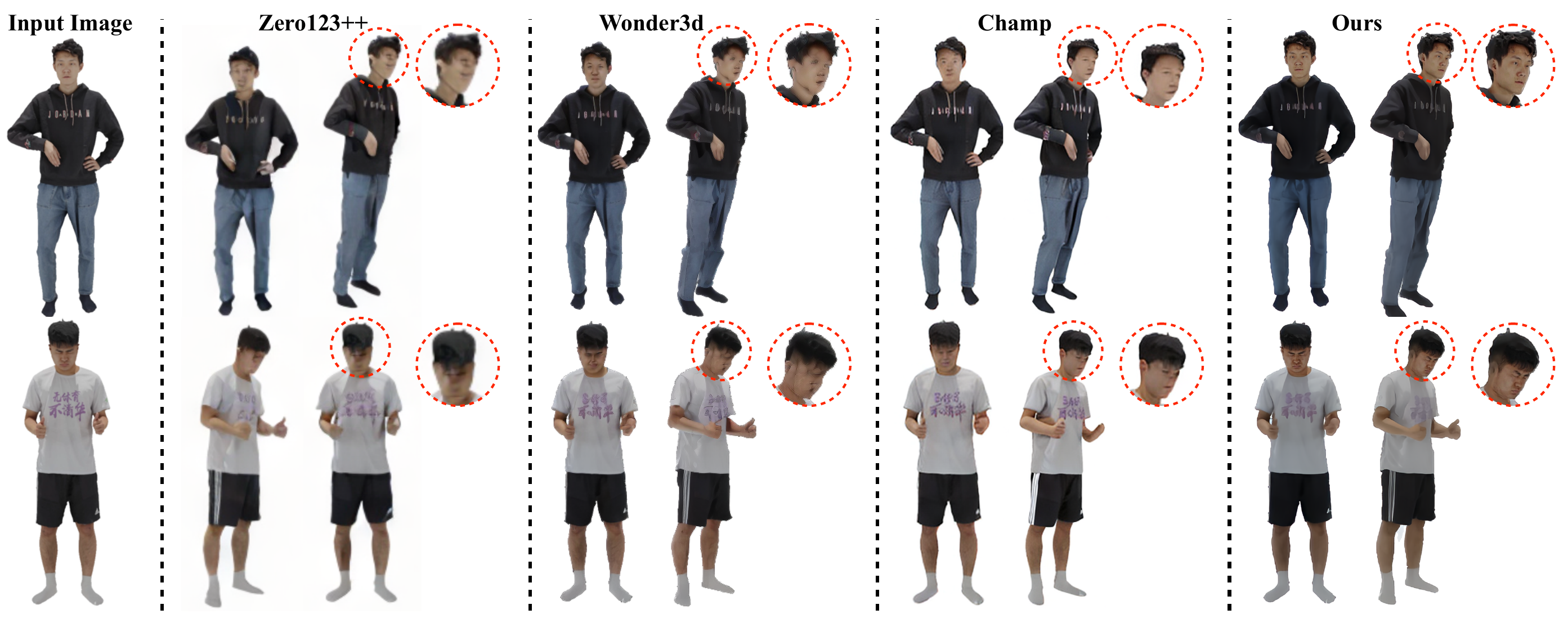}
    %\vspace{-5pt}
    \captionof{figure}{The qualitative comparisons with baseline models on synthesized human multi-view images. Our approach generates superior facial details, producing high-quality and consistent novel human views from a single image input.}
    \label{fig:vis}
    %\vspace{-5pt}
\end{center}%
% \vspace{0.1in}
}]

\begin{abstract}
Generating multi-view human images from a single view is a complex and significant challenge. Although recent advancements in multi-view object generation have shown impressive results with diffusion models, novel view synthesis for humans remains constrained by the limited availability of 3D human datasets. Consequently, many existing models struggle to produce realistic human body shapes or capture fine-grained facial details accurately.
To address these issues, we propose an innovative framework that leverages transferred body and facial representations for multi-view human synthesis. Specifically, we use a single-view model pretrained on a large-scale human dataset to develop a multi-view body representation, aiming to extend the 2D knowledge of the single-view model to a multi-view diffusion model. Additionally, to enhance the model's detail restoration capability, we integrate transferred multimodal facial features into our trained human diffusion model.
Experimental evaluations on benchmark datasets demonstrate that our approach outperforms the current state-of-the-art methods, achieving superior performance in multi-view human synthesis.
\end{abstract}    
\section{Introduction}
\label{sec:intro}

% Talk about Multi-view generation and its problem to be solved
% Multi-view image generation, particularly in relation to humans, has received considerable attention in recent years. This may lead to numerous applications such as AR/VR \cite{Mees2019Selfsupervised3S,Tosun2020RoboticGT}, 3D content creation\cite{Lin2022Magic3DHT}, and virtual try-on \cite{Wang2018TowardCI}. The task setting of multi-view generation is to synthesize novel views from a given input image. Traditional geometry-based approaches attempt to approximate the 3D structure and make pixel-level changes to output new view images \cite{zheng2012smartmanipu,Kholgade20143dmanipulation,kholgade20143dmanipu}; however, these methods often yield only blurry preliminary results. To overcome these limitations, researchers have extensively explored learning-based methods \cite{kalantari2016learnnewview, choi2019extreme, riegler2020freeview, shi2021selfnew, wang2021ibrnet}. While these approaches have achieved some success, ambiguity remains in the synthesized novel views estimated from a single input. To address this issue, a recent study combines global and local features to create a hybrid 3D representation, enabling the network to reconstruct unseen regions without imposing constraints like symmetry, thereby reducing ambiguity and generating expressive details \cite{lin2023vitnerf}.  Subsequent research NerfDiff \cite{gu2023nerfdiff} enhances the quality of unseen features and significantly reduces unclear areas by distilling the knowledge of a 3D-aware conditional diffusion model into neural radiance field. 

Human multi-view image generation has received considerable attention in recent years. This approach enables various applications, including AR/VR \cite{Mees2019Selfsupervised3S,Tosun2020RoboticGT}, 3D content creation \cite{Lin2022Magic3DHT}, and virtual try-on \cite{Wang2018TowardCI}. 
Specifically, the task of human multi-view generation aims to synthesize novel human views from a single input image. The core challenge is to generate consistent multi-view images while minimizing ambiguity as much as possible. 
% Many methods have been proposed to synthesis novel object views to address this challenge. \par
% explain the differentces between geometry and learning based methods. 

Existing methods for multi-view generation can be categorized into either geometry-based approaches \cite{zheng2012smartmanipu,Kholgade20143dmanipulation,kholgade20143dmanipu} or learning-based approaches \cite{kalantari2016learnnewview, choi2019extreme, riegler2020freeview, shi2021selfnew, wang2021ibrnet,Liu2023SyncDreamerGM,Shi2023Zero123AS,liu2024one2345,Long2023Wonder3DSI,Shi2023Zero123AS,Liu2023Zero1to3ZO}. Traditional geometry-based methods aim to generate novel view images by making pixel-level adjustments based on estimated 3D structures \cite{mahajan2009moving}. In contrast, learning-based approaches utilize deep learning techniques to interpret information from a single input. Benefiting from large-scale datasets, many diffusion based methods, such as Zero123++ \cite{Shi2023Zero123AS}, Syncdreamer \cite{Liu2023SyncDreamerGM} and Wonder3D \cite{Long2023Wonder3DSI}, show great performance in generating novel object images with structural coherence.

Though these methods achieve some success for object-oriented tasks, human multi-view generation task is still challenging. 
This is because object-oriented models \cite{Liu2023SyncDreamerGM,Shi2023Zero123AS,liu2024one2345,Long2023Wonder3DSI,Shi2023Zero123AS,Liu2023Zero1to3ZO} benefit from large-scale datasets for training, such as Objaverse \cite{Deitke2022ObjaverseAU, Deitke2023ObjaverseXLAU}. However, due to the inherent limitations of human data, the human body is prone to occlusion in multi-view scenarios.
% In contrast to object-oriented tasks trained on large-scale dataset, such as Objaverse \cite{Deitke2022ObjaverseAU, Deitke2023ObjaverseXLAU}, human multi-view synthesis meets the following problems. 
In other words, large-scale multi-view human data is required to learn human body representations to alleviate this problem. 

Therefore, directly applying existing object-oriented methods to generate novel human views may lead to incomplete or distorted occluded regions. Moreover, human novel view synthesis requires to restore facial details, while these models do not focus on such information, leading to blurred faces and distorted expressions in the results. To summarize, human multi-view generation still suffers from the following problems: 1) how to generate high-quality results with limited scale 3D human datasets and 2) how to capture fine-grained facial details accurately.
%Recently, PanoHead \cite{an2023panohead} designs a 3D-aware GAN to overcome structure ignorance while Preface \cite{buhler2023preface} in$troduce a volumetric human face priors to enhance the generated details. T
 %Convolutional Neural Networks based methods \cite{2016facerecon, liu2017denseface, liu2016jointface, richardson2017learning, zhu2016facealign} and 

% In this part, talk about problems human faces synthesis
% While substantial advancements have been achieved in object synthesis, the performance of human novel view synthesis continues to be unsatisfactory. Insufficient 3D human data directly constrain the capability of current models, thereby making it difficult for them to produce high-quality outcomes. Moreover, 
% help in promoting the appearance with acceptable geometry but also show unreal textures. After that, generative adversarial network (GAN) enables high-resolution synthesis but provides weak inference results in face structures \cite{shen2018faceid}. 

% In this part, talk about 3d reconstruction from single image (connect multiview synthesis with it)
% Merely generating new perspectives of an object is insufficient in numerous scenarios. In Zero123\cite{Liu2023Zero1to3ZO}, researchers try to construct a model that synthesizes novel views by changing camera positions and utilize this prior to reconstruct 3D object.  

% In this part, talk about what we do in this paper
 In this paper, we propose an innovative framework for multi-view human generation through learning transferred face and body representation to address two challenges above. Specifically, to alleviate the problem of scarcity of 3D human datasets, we leverage a pretrained single-view human model to contribute to enrich 3D information to learn transferred body representation. Besides, we employ normal maps from SMPL to model the coarse body shape and pose. It is observed that transferred body representation alone is insufficient to generate detailed and consistent multi-view humans, often missing facial details. However, facial detail restoration is technically challenging, which requires to be structure-accurate and texture-rich. To address this issue, we propose integrating the multimodal information from 2D and 3D facial data to enhance face representation, as the 3D prior contains robust facial structures and the 2D prior provides a person's identity in the semantic space.

% In this part, talk about the main contributions
\par

The main contributions of this paper are summarized as follows:
\begin{itemize}
    \item We leverage a single-view human model, pre-trained on large-scale 2D human data, to learn a transferred multi-view body representation.
    \item We integrate multimodal facial features from 2D and 3D priors to provide structure-accurate and identity-preserving information, enhancing face representation.
    \item Our model demonstrates state-of-the-art performance in generating human multi-views on the THuman2.1 and 2K2K datasets. 
\end{itemize}
\section{Related Work}
\label{sec:rel}
\noindent \textbf{Multi-view Image Generation.}
Multi-view generation has become a significant area of research in recent years, with diffusion-based models leading the way. Numerous studies have focused on synthesizing novel views of objects using these models \cite{Liu2023SyncDreamerGM, liu2024one2345,Long2023Wonder3DSI,Shi2023Zero123AS,Liu2023Zero1to3ZO}. Trained on large-scale 3D datasets \cite{Deitke2023ObjaverseXLAU, Deitke2022ObjaverseAU}, researchers have explored novel view synthesis from a single image. A foundational contribution to this field is Zero123 \cite{Liu2023Zero1to3ZO}, although its outputs often lack consistency across views. Subsequent works, such as Zero123++ \cite{Shi2023Zero123AS} and One2345 \cite{liu2024one2345}, aim to enhance the correlations among predictions and generate more intricate 3D objects. Additionally, SyncDreamer \cite{Liu2023SyncDreamerGM} struggles with geometry and texture quality. Although Wonder3D \cite{Long2023Wonder3DSI} improves texture reconstruction, it still faces challenges with thin structures. Champ \cite{Zhu2024ChampCA} combines SMPL parameters with depth, normal, and semantic maps to enhance visual accuracy but falls short in rendering facial details due to the limitations of the SMPL model. Our method focuses on refining these tiny facial structures.
\newline

\begin{figure*}
\centering
\includegraphics[width=1\linewidth]{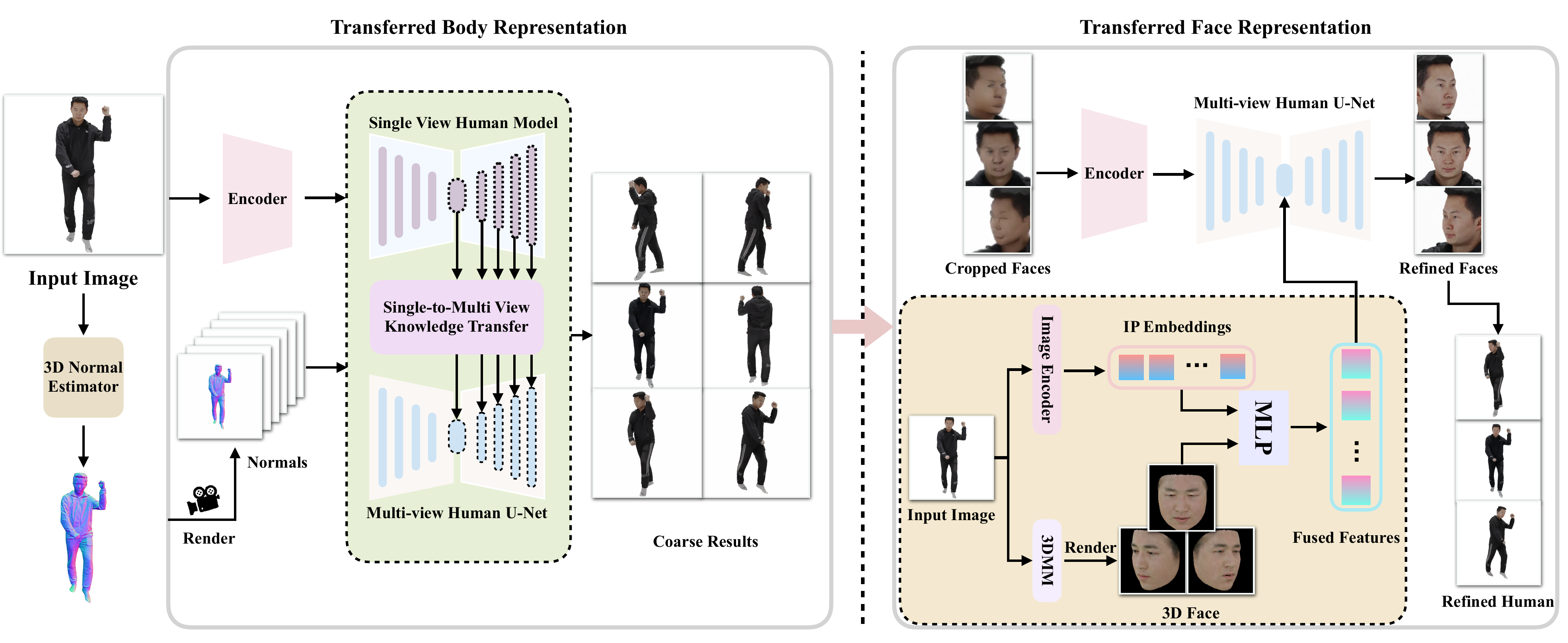}
\caption{\textbf{Overall architecture of our method.} Our method consists of two learning phases. At the the first phase, we leverage a pretrained single-view human \cite{Li2024CosmicManAT} model to learn a transferred body representation. Knowledge transfer is achieved by updating a multi-view human model using pre-trained parameters from a single-view human model. At the second phase, we integrate the 2D and 3D face features to learn a transferred face representation. 3D face features provide structure-accurate priors, while 2D features deliver identity-related priors.}
\label{fig:model}
\end{figure*}

\noindent \textbf{Diffusion Models for Personalization.}
Diffusion models \cite{Ho2020DenoisingDP, ye2023ipadapter, li2024blip} have demonstrated remarkable capabilities in generating personalized human representations, making them widely adopted in the personalized generation process aimed at preserving essential features of specific individuals across different views. Many methods for personalized synthesis primarily focus on accurately capturing the identity of the subject. One notable approach is DreamBooth \cite{Ruiz2022DreamBoothFT}, which fine-tunes a pre-trained text-to-image model to associate a unique identifier with a specific individual. To handle multiple subjects, Custom Diffusion \cite{kumari2023customdiff} utilizes a combination of various fine-tuned models, thereby enhancing personalization capabilities.  Despite the success of these methods, their efficiency often falls short of expectations. To tackle this challenge, HyperDreamBooth \cite{ruiz2024hyperdreambooth} introduces a hypernetwork that efficiently generates a compact set of personalized weights from a single image of an individual, significantly improving inference speed while maintaining high-quality generation. Similarly, PhotoMaker \cite{Li2023PhotoMakerCR} employs a stacked identity embedding as a unified representation, achieving impressive generation results while substantially reducing processing time.
\newline
 
%-------------------------------------------------------------------------
\noindent \textbf{Face Image Restoration.}
Face Restoration has garnered significant interest among researchers \cite{chen2021progressive, tu2021jointface, yue2024difface}. Image restoration can be classified into two categories based on prior information: non-blind restoration \cite{li2023surveyimageres}, which uses known distortion models, and blind restoration \cite{zhu2022blind}, which infers distortion characteristics without prior knowledge. Blind face reconstruction \cite{zhou2022blindface} is a long-term challenging task in high-quality multi-view human synthesis. Some initial solutions, like GFRNet \cite{li2018gfrnet} and GWAINet \cite{dogan2019gwainet} make use of a single exemplar image for restoration, leading to unsatisfactory recovery results when dealing with unfamiliar conditions. ASSFNET \cite{li2020enhanced} suggests using multi-exemplar images and adaptive fusion to enhance the  generalization ability of restoration methods. GFPGAN \cite{wang2021gfpgan} and GPEN \cite{Yang2021GANPE} leverage generative priors as additional guidance, but they also introduce artifacts with severely damaged input. To solve robustness issue, diffusion-based methods \cite{wang2023dr2, ding2024resbygen, lu20243dprior} tend to utilize structure and texture features to constrain the denoising process.  Inspired by the methods above, we leverage 2D and 3D facial features to help our multi-view human UNet capture facial details following the similar refinement process.
\newline

\section{Method}
\label{sec:method}
%-------------------------------------------------------------------------
In this section, we explain the overall framework of our proposed approach, as illustrated in Figure \ref{fig:model}. Given an input human image, the goal of our method is to synthesize novel views with fine-grained and consistent information about the person from different perspectives. In Section \ref{subsec:pre} we present an overview of the latent diffusion model to establish the foundational concepts necessary for the subsequent discussions. Section \ref{subsec:coarse} describes the generation process of the coarse human results, in which 2D human information from a pretrained large-scale model is extended to the multi-view scope to alleviate the problem of limited 3D human data. Additionally, normal maps rendered from SMPL are employed to establish a representation of body shape and pose. In Section \ref{subsec:refine}, we present the process of enhancing facial representation with multimodal facial features fused. 

\subsection{Preliminary}
\label{subsec:pre}
\noindent \textbf{Diffusion Models.} Our method is based on Diffusion Model \cite{Ho2020DenoisingDP, SohlDickstein2015DeepUL, Song2020ScoreBasedGM}, a class of promising generative models that introduce noise to data and then reverse this process to generate data from the noise. This procedure, known as the forward process or noise-infection procedure, is formalized as a Markov chain: $q(X^{1},...,X^{T}|X^{0})=\prod_{t=1}^{T}{q(X^{t}|X^{t-1})}$.
It adds small Gaussian noise to latent images at the previous time step as follows: $q(X^{t}|X^{t-1})=\mathcal{N}(X^{t};{\sqrt{1-\beta_{t}}}{X^{t-1}}, \beta_{t}\boldsymbol{I}),$ where $\beta_{t}$ is a small positive constant and $\boldsymbol{I}$ is the identity matrix with the same dimension as $X^{t-1}$.
\par

The reverse process is defined as the conditional distribution $p_\theta(X^{(0:T-1)}|X^{T}, c)$, where $c$ is the condition guidance. The entire reverse process can be factorized into multiple transitions based on the Markov chain:
\begin{equation}
  p_\theta{(X^{0},...,X^{T-1}|X^{T},c)}=\prod^{T}_{t=1}{p_\theta(X^{t-1}|X^{t},c)}.
  \label{eq:reverse_trans}
\end{equation}\noindent
In the reverse process, the Markov kernel is parameterized as: $p_\theta(X^{t-1}|X^{t},c)=\mathcal{N}({X^{t-1};\mu_\theta{(X^t,t),\sigma_{\theta}^2{(X^t,t)}\boldsymbol{I}})}$,
where $\mu_\theta{(X^t,t)}$ and $\sigma_{\theta}^2{(X^t,t)}$ are mean and variance of the reverse process respectively. By utilizing the reverse transitions $p_\theta{(X^{t-1} | X^{t},c)}$, the latent variables are gradually returned to the images that aligns with the diffusion time-step and image condition.
\par

For training objective, we optimize the simple objective proposed by DDPM \cite{Ho2020DenoisingDP}:
\begin{equation}
\begin{aligned}
    \underset{\theta}{\text{min}}L(\theta) := \underset{\theta}{\min}\mathbb{E}||\epsilon -\epsilon_{\theta}(\sqrt{\Bar{\alpha}_t}X^0+\sqrt{1-\Bar{\alpha}_t}\epsilon,t,c)||^{2}.
  \label{eq:trian_loss}
\end{aligned}
\end{equation}

\subsection{Transferred Body Representation}
\label{subsec:coarse}
In this module, we consider how to generate consistent multi-view images with only limited 3D human data available. The overall framework is shown on the left side of the Figure \ref{fig:model}. Specifically, The initial input to network in the stage consists of two types of images: human image and SMPL normal maps. SMPL is estimated by the input image and then rendered to different views to generate corresponding normal maps. Since normal maps contain limited fine-grained information, they can only be used to guarantee the coarse detail such as body shape and pose.
\subsubsection{Single-view to Multi-view Transfer Module}
This module mainly consists of two UNets, aiming to adopt 2D human features to benefit the multi-view generation process and alleviate the problem of limited 3D human data. The core objective of our module is to represent the input image with detailed body information that can be injected into our multi-view human UNet. Trained on a large-scale dataset CosmicMan-HQ with 6 million high-quality real-world human images, CosmicMan \cite{Li2024CosmicManAT} provides rich human-related 2D information that can be transferred to multi-view generation process. We adopt a framework identical to the denoising UNet, inheriting weights from the CosmicMan, with weight updates conducted independently for each. During training, the input image is first encoded by a VAE encoder. The VAE embeddings are then fed into the single-view human UNet, termed as single-Net. Encoded by the single-Net, appearance information in the input image can be fully parsed. Specifically, appearance information is represented by a set of normalized attention hidden states for the middle and upsampling blocks of the UNet and then is written into the memory box. Our multi-view human UNet, termed as multi-UNet is implemented based on Wonder3D \cite{Long2023Wonder3DSI}, which introduces a cross-domain attention mechanism to ensure consistency between the generated images.  The multi-UNet reads stored features from the memory box, transferring 2D knowledge to the multi-view domain. These features are passed to the spatial self-attention layers in the multi-UNet blocks by concatenating each feature with the original UNet self-attention hidden states to inject the appearance information.

\subsubsection{Normal Map Guidance}
With only one image as input, it is insufficient to represent the 3D geometry of the human due to the occlusion and view-dependent effects. Therefore, additional guidance introduced to the model is necessary. As a 3D parametric human model, SMPL offers a unified representation that encompasses both shape and pose. Given an input image of a human body,  we can adopt the 4D-Humans \cite{Goel2023HumansI4} to estimate the parameters of SMPL. We can model the coarse body shape and pose from the different target perspective by rendering the SMPL mesh to obtain 2D representations. Specifically, we render normal maps from SMPL with the target poses to guide multi-view human UNet in generating human novel views. All encoded normal maps serve as additional conditions guiding the generation process.
\begin{figure}
\centering
\includegraphics[width=1\linewidth]{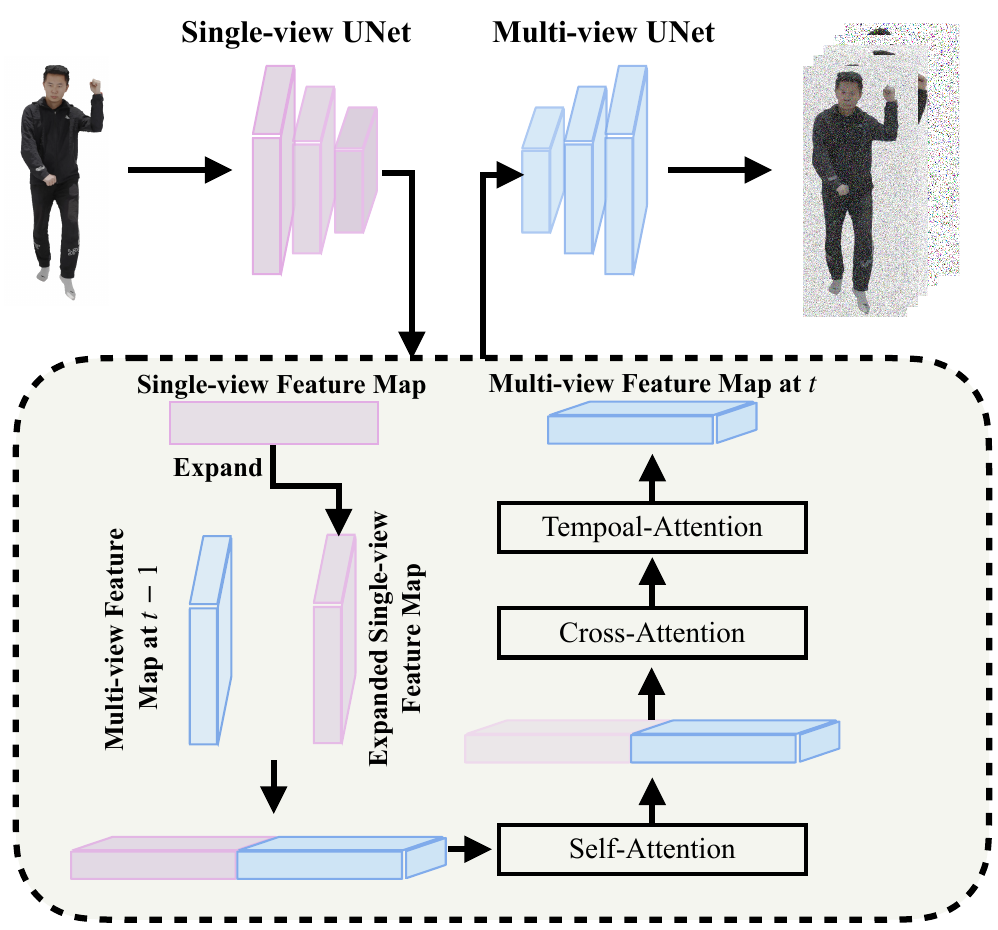}
\caption{ Illustration of single-to-multi view knowledge transfer. Single-view feature map from input image is expanded to the same dimension with multi-view feature map from previous timestep $t-1$. Then multi-view feature map at timestep $t$ is obtained through attention operation.}
\label{fig:knowledge_transfer}
\end{figure}

\subsection{Transferred Face Representation}
\label{subsec:refine}
Since a single image and SMPL normal maps cannot provide all the information needed for generating high-quality novel views, the results from previous module often fail to reconstruct some details, such as human face. To address this problem, we propose integrating 2D and 3D facial features to learn a face representation. Specifically, we regard the refinement process as a face restoration problem and propose a novel framework to solve this problem. To leverage pretrained information and keep the architecture neat and powerful, we adopt the pretrained multi-UNet as the backbone. There are two main differences between these two stages. On the one hand, the condition of the model only contains face images from different views. On the other hand, we propose a novel facial feature enhancement module to increase the performance of face refinement. The implementation details are as follows.
\subsubsection{Face Segmentation and Restoration}
To align with the face restoration task, we first identify the face region by employing an existing framework, RetinaFace \cite{serengil2024lightface}, which allows us to accurately detect and crop the facial area from the input image. Then, we apply a super-resolution model to enhance the resolution of the cropped face images, scaling them to match the resolution of the input image. This consistency in resolution is crucial, as it enables the model to capture finer facial details and produce high-quality results. Following the approach outlined in GPEN \cite{Yang2021GANPE}, we then restore the refined face images back into the original image context using the mask corresponding to the cropped face region. This ensures that the enhanced facial details are seamlessly integrated into the overall image, preserving the integrity of the original scene while improving the quality of the facial features. The process allows for effective face restoration, enhancing both the fidelity and realism of the synthesized faces in multi-view human synthesis tasks. 

\subsubsection{Face Representation Learning}
To gain fine-grained facial detail, it is not sufficient to rely solely on the guide image from the previous module. 2D and 3D priors are complementary to each other. In this paper, differing from existing face restoration methods \cite{zhu2022blind,lu20243dprior,Chen2024BlindFR}, we first propose integrating the 2D and 3D facial features into the 3D human UNet to ensure the fidelity, authenticity, and identity consistency. We follow \cite{Li2023PhotoMakerCR} to extract ID-related embeddings from the input image as the 2D prior, providing a representation of the person's ID in the semantic space. Compared to 2D priors, 3D priors are useful in providing robust facial structures, which is essential for further restoration of fine details. Same with \cite{zhu2022blind}, we follow the practice the D3DFR \cite{Deng2019Accurate3F} to predict the coefficients of 3D morphable face models (3DMMs \cite{Blanz1999AMM}). With predicted 3DMM coefficients, the 3D face shape $\hat{S}$ and albedo texture $\hat{T}$ are presented as:
\begin{align}
    &\hat{S} = \hat{S}(\alpha,\beta)=\bar{S}+B_{id}\alpha+B_{exp}\beta ,\\
    &\hat{T}=\hat{T}(\delta)=\bar{T}+B_{t}\delta,
\end{align}
where $\alpha$, $\beta$ and $\delta$  are the coefficients of the Basel Face Model (BFM) for identity, expression and BFM texture respectively. $\bar{S}$ and $\bar{T}$ are the mean face shape and albedo texture. $B_{id}$, $B_{exp}$ and $B_t$ denote the PCA bases of identity, expression and texture, illumination and face pose. Estimated with 3DMM, it is easy to project the 3D face onto 2D images according to the given poses based on a differentiable mesh renderer. The process of rendering can be represented as:
\begin{equation}
    I_{3d} = \mathcal{F}_{render}(\hat{S},p),
\end{equation}
where $p$ is the given pose. Subsequently, the rendering facial 3D images $I_{3d}^{i}$ are encoded by ResBlock, same as SR3 \cite{Saharia2021ImageSV}. This process can be expressed as:
\begin{equation}
    \mathcal{F}_{3d}^{i} = ResBlock(I_{3d}).
\end{equation}
In order to  take advantage of the strengths of the 2D and 3D priors and obtain a better representation of the face features, we propose integrating 2D and 3D features in the semantic space. We denote IP embedding as $\mathcal{F}_{2d}$. Inspired by PhotoMaker \cite{Li2023PhotoMakerCR}, we use two MLP layers to fuse the $\mathcal{F}_{2d}$ and $\mathcal{F}_{3d}$. Finally, the combined features, denoted as $\mathcal{F}_{align}$, are fed into the multi-level feature extraction module in 3D human UNet to extract facial details, thereby capturing both structural and identity information within the priors.

\subsection{Training and Inference Details}
\noindent \textbf{Training.}
Our method consists of two distinct stages. In the first stage, our method aims to learn a body representation leveraging pretrained sing-view human model. In this phase, we jointly optimize the parameters of the single view human model and multi-view human UNet, while maintaining the weights of the VAE encoder and decoder in a frozen state, as well as the CLIP image encoder. The training data for this stage consists of pairs of images from different viewpoints. The input image is the front view of the character, and the target image consists of six different perspectives. Normal maps are rendered from the estimated SMPL model based on target poses. For the transferred face representation learning, we train to update the weights of the multi-view human UNet and the face embedding module. We extract the IP embedding and 3D face features from the input image to enhance face representations.\par

\noindent \textbf{Inference.}
During the inference process, novel-view human synthesis is performed on a single image by transferring knowledge to build superior body and face representations. The transferred body representation is utilized to provide a base body shape for generation. Additionally, the transferred face representation provide structure-accurate 3D priors and IP-rich 2D priors to help restore facial details. Both representations serve as powerful means to synthesis high-quality and consistent human novel views with only one image input and limited training data available.

\begin{figure*}
\centering
\includegraphics[width=1\linewidth]{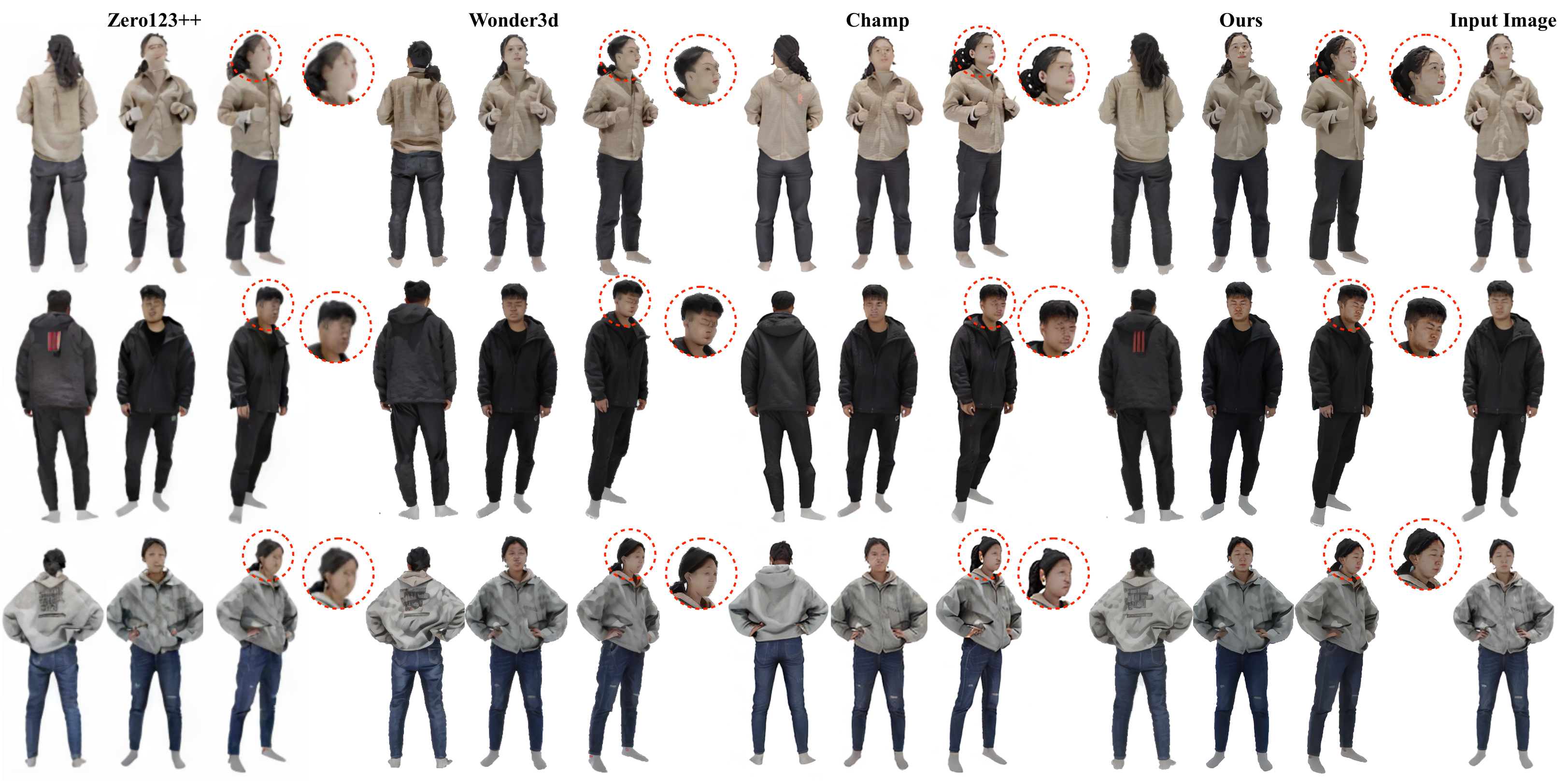}
\caption{Qualitative comparisons between our and the state-of-the-art approaches on the THuman2.1 dataset are presented. The red circle highlights the side-view face area in the synthesized multi-view, where our method demonstrates superior face detail synthesis.}
\label{fig:vis_compare}
\end{figure*}
\section{Experiments}
\label{sec:exp}

In this section, we assess the effectiveness of our method by evaluating the novel-view human synthesis task. Specifically, we compare our method with previous state-of-the-art models through quantitative and qualitative comparisons. To further analyze its modeling capability, we directly extend our method trained on the THuman2.1 to the 2K2K dataset \cite{han2023high}, enabling a more detailed and robust assessment. Additionally, we perform ablation studies to investigate the individual contributions and overall impact of each component in our framework, offering deeper insights into its effectiveness for this task.

\subsection{Setup}
\noindent \textbf{Implementation Details.} To generate high-quality and consistent novel views, our method is implemented on the backbone of Wonder3D \cite{Long2023Wonder3DSI}. The overall framework is optimized using Adam \cite{Kingma2014AdamAM} on 1 NVIDIA A800 GPU for about 4 days with a batch size of 4. We set the learning rage as $1e -4$ for the trainable modules.  The training regimen is structured in two phases. For the first phase, we train the 3D human UNet to generate 6 views of full human body. In the second phase, we train the module to refine the face area using the cropped faces input from the first phase. It's worth noting that in certain views, such as back view, face area cannot be detected. For a uniform training data format, we choose only three of these perspectives including the front view, the front left view and the front right view, for refinement. For both phases, the resolution of the training data is resized to 512 $\times$ 512. During the inference stage, we perform 50 steps of the DDIM sampler \cite{Song2020DenoisingDI} and the scale of the classifier-free guidance is set to 3.\par

\begin{figure*}
\centering
\includegraphics[width=0.9\linewidth]{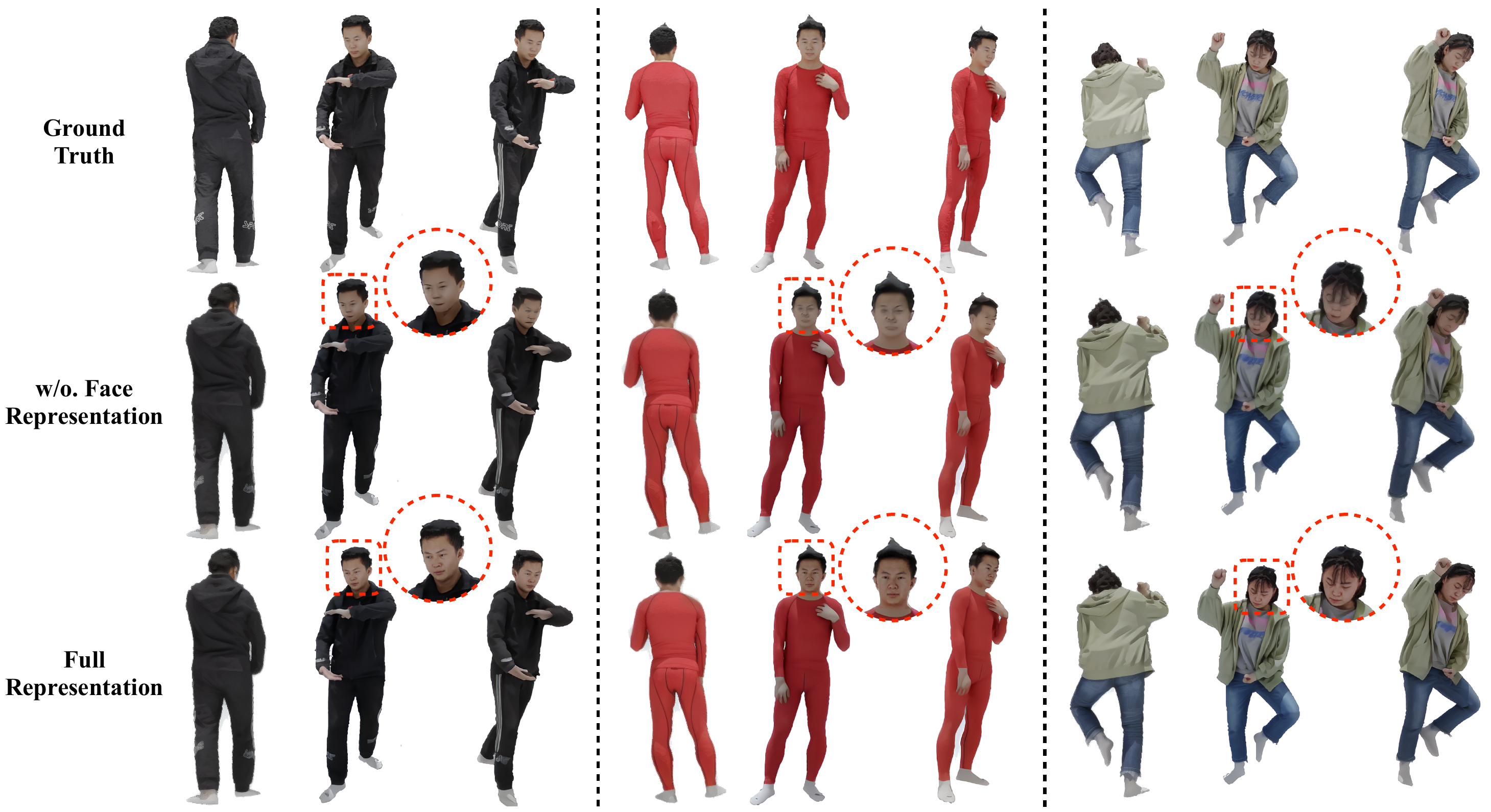}
\caption{Qualitative results of the two transferred representation learning phases. The first row shows the ground truth, the second row displays results from the transferred body representation learning, and the third row shows results with the full representation.}
\label{fig:abl_stage}
\end{figure*}

\noindent \textbf{Evaluation metrics.} To demonstrate the superiority of our approach, we evaluate its performance using the established metric utilized in existing research. Specifically, we assess the novel-view human synthesis quality with three widely used metrics: Peak Signal-to-Noise Ratio (PSNR) \cite{Hor2010ImageQM}, Structural Similarity Index (SSIM) \cite{Wang2004ImageQA} and Learned Perceptual Image Patch Similarity (LPIPS) \cite{Zhang2018TheUE}. PSNR is widely used to measure the fidelity of the reconstruction in terms of pixel-level accuracy. SSIM evaluates the perceptual similarity between images by considering luminance, contrast and structure, thereby offering insights into the perceived quality of the synthesized images. LPIPS is a perceptually aligned metric that utilizes deep network features to capture high-level similarities and better correlates with human judgments on image similarity.\par

\begin{table}
\centering
\resizebox{0.85\linewidth}{!}{
\begin{tabular}{l|ccc}
\toprule
   \textbf{Methods} & PSNR$\uparrow$ & SSIM$\uparrow$ & LPIPS$\downarrow$ \\
\midrule
      Syncdreamer \cite{Liu2023SyncDreamerGM} &13.24  &0.811 &0.209\\
      One2345 \cite{liu2024one2345} &15.53&0.833 &0.215\\
\midrule
      Zero123++ \cite{Shi2023Zero123AS} $^{\dag}$ &20.33 &0.874 &0.118\\
      Wonder3D \cite{Long2023Wonder3DSI}$^{\dag}$ &23.42 &0.906 &0.111\\
      Champ \cite{Zhu2024ChampCA} $^{\dag}$ &25.26 &0.942 &0.063\\
\midrule
      Ours&\textbf{27.13} &\textbf{0.986} &\textbf{0.022}\\
\bottomrule
\end{tabular}}
\caption{We present the quantitative metrics of all methods on the THuman2.1 dataset in terms of PSNR, SSIM and LPIPS. Methods finetuned on the THuman2.1 dataset are marked with $^\dag$.}
\label{tab:exp_metrics}
\end{table}

\subsection{Comparisons}
\noindent \textbf{Dataset.} 
We conduct main experiments on widely used 3D human datasets, called THuman2.1, which includes approximately 2500 high-quality human scans \cite{Yu2021Function4DRH}. Specifically, our training dataset comprises 2350 scans from THuman2.1 and the rest for validation. To demonstrate the generalization ability of our method, we directly apply our pretrained model to another common 3D human dataset 2K2K \cite{han2023high}. It is a large-scale dataset containing high-quality  3D human model of 2,050 individuals.\par
\noindent \textbf{Baselines.} We conducted a comprehensive evaluation of our method against state-of-the-art novel-view human synthesis approaches, including SyncDreamer \cite{Liu2023SyncDreamerGM}, One2345 \cite{liu2024one2345}, Zero123++ \cite{Shi2023Zero123AS}, Wonder3D \cite{Long2023Wonder3DSI}, and Champ \cite{Zhu2024ChampCA}. Among these, SyncDreamer, One2345, Zero123++, and Wonder3D have been demonstrated to be effective for multi-view object synthesis tasks, making them suitable benchmarks for comparison in our study. For our experiments, we fine-tuned the Zero123++ and Wonder3D models on the THuman2.1 dataset using their publicly available implementations to ensure consistency and fairness in evaluation. Since Champ is originally designed for human animation tasks using reference videos, we adapted its framework for the novel-view human synthesis task by modifying the input and output processing stages to align with the requirements of our evaluation. This enabled us to assess Champ's potential in the context of novel-view synthesis, despite its primary focus on animation-based tasks. \par
\noindent \textbf{Evaluation on THuman2.1 Datasets.} The qualitative results of novel-view human synthesis are displayed in Figure \ref{fig:vis_compare}. As shown in the results, most of the baselines can capture semantic information from the single input image to generate plausible novel views. However, most of them fail to reconstruct facial details. In comparsion, our method can restore more realistic face from the single image. Table \ref{tab:exp_metrics} presents the quantitative results of our method and the baselines on the THuman2.1 dataset, focusing on commonly used metrics such as PSNR, SSIM and LPIPS. We generate six views of human from different perspective when calculating the related metrics. Table \ref{tab:exp_metrics} shows that our method outperforms other baselines achieving higher PSNR and SSIM scores, and lower LPIPS values. In conclusion, both qualitative and quantitative results demonstrate the superior performance of our method, which achieves state-of-the-art performance in the multi-view human synthesis task.\par
\begin{figure}
\centering
\includegraphics[width=1\linewidth]{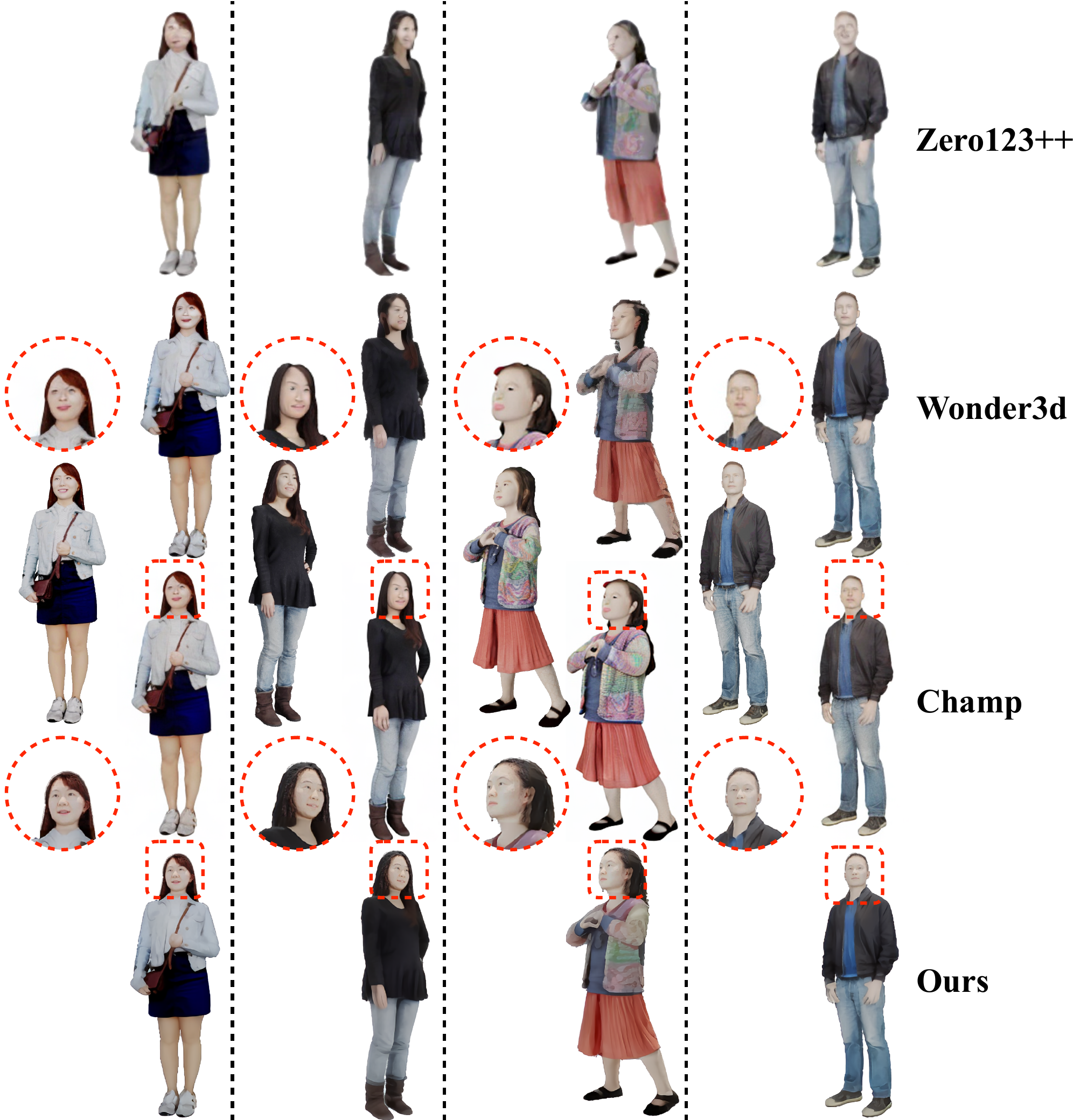}
\caption{Qualitative results of our method and baselines on the 2K2K dataset. The red circle highlights the face area to facilitate comparison of our generation results with those of the baselines.}
\label{fig:vis_2k2k}
\end{figure}

\noindent \textbf{Evaluation on 2K2K Datasets.} To evaluate the generalization capability of our model, we directly conduct experiments on the 2K2K \cite{han2023high} dataset using pretrained model from THuman2.1. From Figure \ref{fig:vis_2k2k}, we can see that our approach achieved satisfactory performance on this dataset, demonstrating its robustness and adaptability to new data distributions beyond the training set. Notably, compared to other baselines the model maintained high accuracy and visual consistency across multiple views, reflecting its strong capacity to capture complex human structures and details under diverse conditions. These results underscore the model’s generalization ability and superior performance.\par

\subsection{Ablation Studies}
\noindent \textbf{Transferred Body Representation.} To demonstrate the effectiveness of our transferred body representation learning design, we conduct ablation studies by training our model to generate coarse outputs both with or without the knowledge transfer module and normal guidance. As illustrated in Table \ref{tab:exp_abl_body}, removing both modules results in reduced performance across all three metrics, thereby validating the effectiveness of our body representation design.\par

\begin{table}
\centering
\resizebox{0.9\linewidth}{!}{
\begin{tabular}{l|ccc}
\toprule
   \textbf{Methods} & PSNR$\uparrow$ & SSIM$\uparrow$ & LPIPS$\downarrow$ \\
\midrule
    Ours (Full) &\textbf{27.130} &\textbf{0.986} &\textbf{0.022}\\
\midrule
    w/o. Knowledge Transfer  &24.613  &0.951 &0.048 \\
    w/o. Normal Guidance &26.023 &0.945 &0.038 \\
\bottomrule
\end{tabular}}
\caption{Ablation study for the transferred body representation learning stage. Without knowledge transfer module or normal guidance may lead to poorer performance.}
\label{tab:exp_abl_body}
\end{table}

\noindent \textbf{Transferred Face Representation.} To assess the impact of the face enhancement module, as illustrated in the bottom-right section of the Figure \ref{fig:model}, we train the model with or without this module to refine the cropped faces. Based on the visualization results presented in Figure , we can conclude that without the face refinement module, the synthesized face struggles to preserve identity-preserving (IP) features and lacks detailed facial reconstruction.\par

\noindent \textbf{Representation Stage.} To provide a comprehensive comparison of the two stages in our representation learning framework, we present visual results in Figure \ref{fig:abl_stage}. These results demonstrate that the body representation learning stage successfully produces relatively plausible and coherent multi-view human body reconstructions, capturing the overall structure and form across different perspectives. However, the results from this stage fail to reconstruct facial detail. By incorporating the face representation stage, our method achieves significantly improved realism with more accurate facial details, enhancing the overall quality of the synthesized human representations. This two-stage approach thereby ensures both structural coherence and high-fidelity detail, particularly in critical areas like the face.\par

\begin{figure}
\centering
\includegraphics[width=1\linewidth]{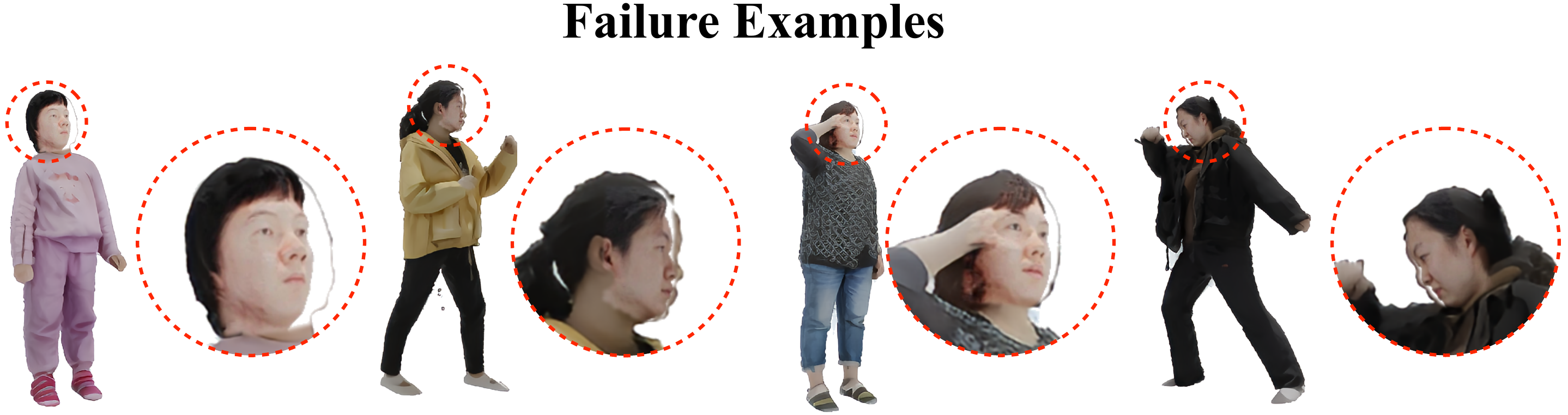}
\caption{Visualization results of failure cases on the THuman2.1 and 2K2K datasets. In these cases, the model fails to reconstruct faces accurately, resulting in discontinuities in the face area.}
\label{fig:failure_vis}
\end{figure}

\noindent \textbf{Limitations.}
There are certain limitations to our model. One key challenge lies in accurately detecting the correct facial region from multiple perspectives, especially when occlusion occurs due to the character's pose. Therefore, our method cannot be adopted to enhance the facial area in all cases. On the other hand, potential mismatches between the refined face and the extracted face region may lead to inconsistencies. In Figure \ref{fig:failure_vis}, we show some failure cases where incomplete restoration occurs.\par

\begin{figure}
\centering
\includegraphics[width=1\linewidth]{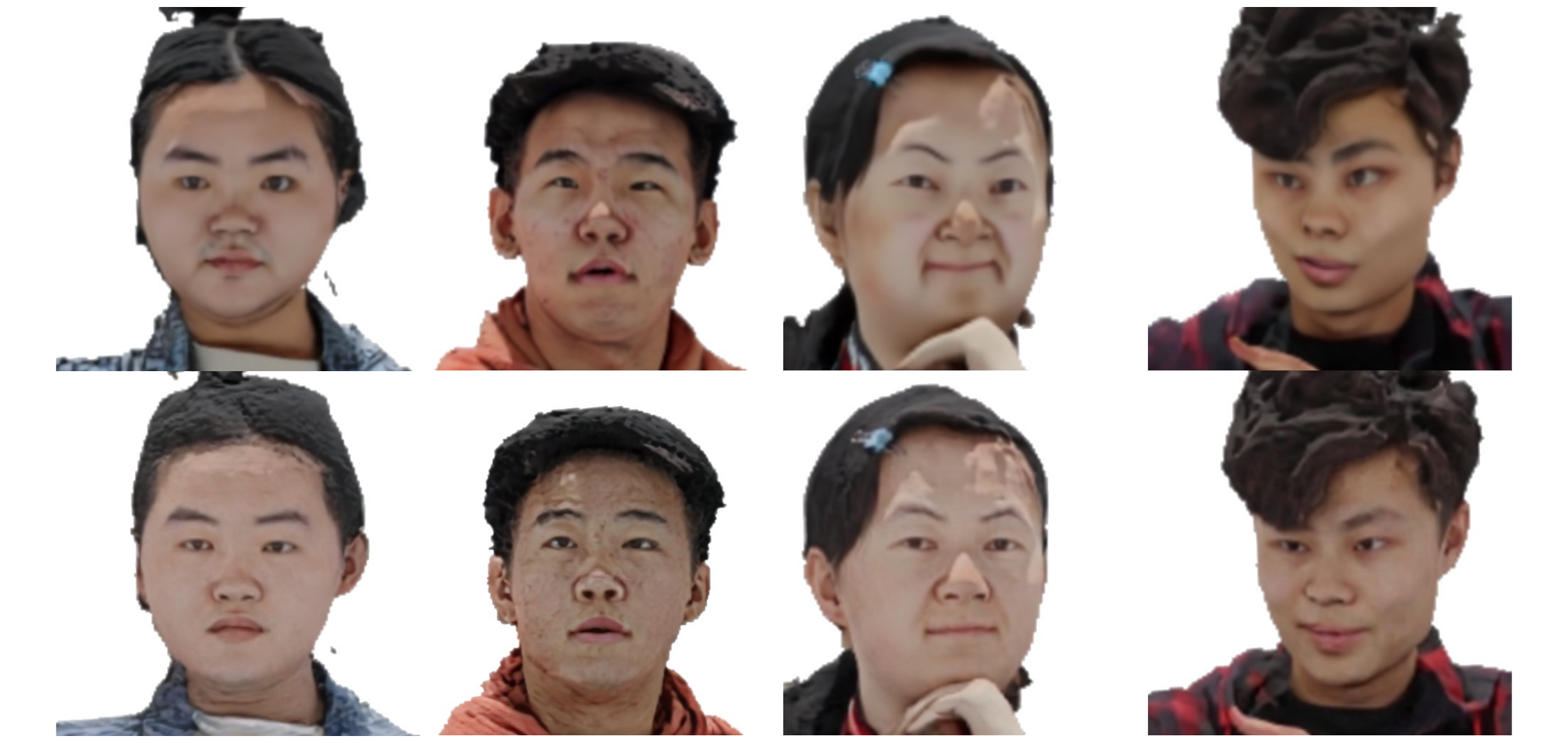}
\caption{Ablation study on the effectiveness of the face embedding fusion module. The first row indicates that the model does not contain a face embedding fusion module and the second row represents that the model maintains a complete structure. }
\label{fig:abl_face}
\end{figure}

\section{Conclusion}
\label{sec:conc}
In this paper, we first present a novel framework based on a diffusion model for human multi-view generation task. Compared with the previous methods, our method extends 2D information from large-scale human datasets to the multi-view scope to address the problem of limited human data. To regain high-quality faces, we incorporate 2D and 3D facial embeddings into the latent diffusion to refine the facial details. We conduct extensive experiments on custom human datasets, THuman2.1 and 2K2K, to show that our method can generate multi-view consistent and high quality results, achieving state-of-the-art performance. 
{
    \small
    \bibliographystyle{ieeenat_fullname}
    \bibliography{main}
}


\end{document}